\documentclass[letterpaper, 10 pt, journal]{IEEEtran}  

\pdfoutput=1
\usepackage{CJKutf8}
\usepackage{color}
\newcommand{\ie}{i.e.,\ }

\newcommand{\Reffig}[1]{Figure~\ref{#1}}
\newcommand{\Refsec}[1]{Section~\ref{#1}}

\newcommand{\Refeq}[1]{Equation~\ref{#1}}
\newcommand{\Reftab}[1]{Table~\ref{#1}}

\usepackage{graphics} 
\usepackage{epsfig} 
\usepackage{algorithm}
\usepackage{algpseudocode}
\usepackage{amsmath}
\usepackage{amssymb}
\usepackage{subfigure}
\usepackage{graphicx}
\usepackage{subfigure}
\usepackage{booktabs}
\usepackage{threeparttable}
\usepackage{multirow}
\usepackage{multicol}
\usepackage{makecell}
\usepackage{dcolumn}
\usepackage[T1]{fontenc}
\usepackage{float}
\newcolumntype{d}[1]{D{.}{.}{#1}}

\usepackage{changes}

\IEEEoverridecommandlockouts
\begin{document}

\title{VNI-Net: Vector Neurons-based Rotation-Invariant Descriptor for LiDAR Place Recognition}
\author{Gengxuan Tian$^{1}$, Junqiao Zhao$^{*,1 , 2}$, Yingfeng Cai$^{1}$, Fenglin Zhang$^{1}$, Wenjie Mu$^{1}$, Chen Ye$^{1}$
    \thanks{This work is supported by the National Key Research and Development Program of China (No. 2021YFB2501104). \emph{(Corresponding Author: Junqiao Zhao.)}}
    \thanks{$^{1}$All authors are with Department of Computer Science and Technology,
        School of Electronics and Information Engineering, Tongji University, Shanghai, China, and the MOE Key Lab of Embedded System and Service Computing, Tongji University, Shanghai, China
            {\tt\footnotesize (Corresponding Author: zhaojunqiao@tongji.edu.cn)}}
    \thanks{$^{2}$Institute of Intelligent Vehicles, Tongji University, Shanghai, China}
}
\maketitle

\begin{abstract}

    LiDAR-based place recognition plays a crucial role in Simultaneous Localization and Mapping (SLAM) and LiDAR localization.
    Despite the emergence of various deep learning-based and hand-crafting-based methods, rotation-induced place recognition failure remains a critical challenge.
    Existing studies address this limitation through specific training strategies or network structures.
    However, the former does not produce satisfactory results, while the latter focuses mainly on the reduced problem of $\mathbb{SO}(2)$ rotation invariance. Methods targeting $\mathbb{SO}(3)$ rotation invariance suffer from limitations in discrimination capability.
    In this paper, we propose a new method that employs Vector Neurons Network (VNN) to achieve $\mathbb{SO}(3)$ rotation invariance.
    We first extract rotation-equivariant features from neighboring points and map low-dimensional features to a high-dimensional space through VNN.
    Afterwards, we calculate the Euclidean and Cosine distance in the rotation-equivariant feature space as rotation-invariant feature descriptors.
    Finally, we aggregate the features using GeM pooling to obtain global descriptors.
    To address the significant information loss when formulating rotation-invariant descriptors, we propose computing distances between features at different layers within the Euclidean space neighborhood.
    This greatly improves the discriminability of the point cloud descriptors while ensuring computational efficiency.
    Experimental results on public datasets show that our approach significantly outperforms other baseline methods implementing rotation invariance, while achieving comparable results with current state-of-the-art place recognition methods that do not consider rotation issues.

\end{abstract}



\section{Introduction}

\IEEEPARstart{P}{lace} recognition plays a crucial role in both Simultaneous Localization and Mapping (SLAM) and localization systems.
It serves to eliminate cumulative errors in SLAM by providing loop closures and offers robust and efficient localization when odometry fails.
In applications such as autonomous driving, LiDAR-based place recognition methods  \cite{zhangPCAN3DAttention,uyPointNetVLADDeepPoint2018,liuLPDNet3DPoint2019} have gained preference over vision-based methods  \cite{arandjelovicNetVLADCNNArchitecture} due to their higher accuracy and less susceptibility to illumination and weather variations.

Numerous LiDAR-based place recognition methods have been proposed and have achieved promising results.
Some methods rely on hand-crafted global descriptors to achieve rapid localization  \cite{heM2DPNovel3D2016,kim1DayLearning1Year2019,joestarkLiDARIrisLoopClosure2022,xuRINGRototranslationInvariant2022}.
However, such methods exhibit limited scene description and generalization abilities in large-scale place recognition tasks.
Others extract point cloud features based on deep learning and apply these features to place recognition tasks.
PointNetVLAD  \cite{uyPointNetVLADDeepPoint2018} represents the first large-scale LiDAR-based place recognition solution based on deep learning.
It employs PointNet  \cite{charlesPointNetDeepLearning2017} for feature extraction and NetVLAD  \cite{arandjelovicNetVLADCNNArchitecture} for aggregating local features to obtain global descriptors.
Subsequent methods improve the discriminability of global descriptors by incorporating hand-crafted features  \cite{liuLPDNet3DPoint2019}, attention mechanisms  \cite{huiPyramidPointCloud,zhangPCAN3DAttention,fanSVTNetSuperLightWeight2021}, or sparse convolution  \cite{komorowskiMinkLocLidarMonocular2021,xuTransLoc3DPointCloud}.
Despite the notable advancements, these methods
cannot effectively address the problem of rotation-induced place recognition failure.
In autonomous driving scenarios, the pre-established map cannot cover all orientations of the same place.
As a result, when a vehicle travels to the same place with different orientations, the differences between the extracted global descriptor and the descriptors in the map can be significant, which substantially reduces the robustness of localization.

Existing place recognition methods for solving the rotation problem can be categorized into two groups: training strategies-based and network architecture-based.
In the first category, techniques such as data augmentation and registration auxiliary tasks are adopted.
Data augmentation-based methods  \cite{zhangPCAN3DAttention,uyPointNetVLADDeepPoint2018,liuLPDNet3DPoint2019} improve the robustness of global descriptors to rotation changes by randomly rotating point clouds during training.
Registration auxiliary task-based methods  \cite{komorowskiEgoNNEgocentricNeural2022,vidanapathiranaLoGG3DNetLocallyGuided2022,cattaneoLCDNetDeepLoop2022} optimize the global descriptor's rotation robustness by coupling the registration-based metric localization with the place recognition task.
Although both methods improve the robustness of global descriptors to rotation, they increase the complexity of network training.

The second category introduces specific network architecture design to address the rotation problem.
These include translation-equivariant-based, and rotation-equivariant neural network-based methods.
Translation-equivariant-based methods  \cite{liRINetEfficient3D2022,maOverlapTransformerEfficientRotationInvariant2022,xuDiSCODifferentiableScan2021,maCVTNetCrossViewTransformer2023} project 3D point clouds onto a 2D plane to convert 3D rotations into 2D translations.
However, the projection sacrifices the 3D information of the point cloud and only provides rotational robustness in $\mathbb{SO}(2)$, \ie the Yaw angle.
To address the rotational robustness problem in $\mathbb{SO}(3)$, researchers propose using rotation-equivariant neural networks to directly extract rotation-invariant descriptors from 3D point clouds for place recognition tasks.
SphereVald  \cite{zhaoSphereVLADAttentionbasedSignalenhanced2022} applies spherical convolution  \cite{estevesLearningEquivariantRepresentations2018} to extract rotation-equivariant features.
$\mathbb{SE}(3)$-Equivariant  \cite{linSEEquivariantPoint} introduces group convolution and uses GeM Pooling \cite{radenovicFineTuningCNNImage2019} to aggregate local features to obtain global descriptors.
RPR-Net  \cite{fanRPRNetPointCloudbased2022} constructs multiple rotation-invariant features directly from the original point cloud and extracts global descriptors using a dense network architecture and GeM Pooling\cite{radenovicFineTuningCNNImage2019}.
However, these methods have all to some extent reduced the representational capability of descriptors for the scene.

In this paper, we propose to extract rotation-invariant and discriminative 3D point cloud global descriptors using Vector Neurons Networks (VNN)  \cite{dengVectorNeuronsGeneral2021}.
We construct rotation-equivariant features for the local neighborhood of the point cloud and obtain rotation-equivariant high-dimensional space descriptors through VNN encoding.
The original VNN learns an orientation for the entire point cloud using Multi-Layer Perceptron (MLP) and aligns the point cloud based on the estimated orientation to achieve rotation invariance.
However, this is limited to 3D object classification tasks where the ground truth orientation is assumed to be known.

To address this limitation, inspired by RIConv  \cite{zhangRotationInvariantConvolutions2019} which achieves rotation-invariant by computing relative distances, we propose the Euclidean and Cosine distance-based layers to achieve rotation invariance in the feature space.
The Euclidean distance layer calculates the absolute distance between each point in different feature spaces, while the Cosine distance layer first incorporates a rotation-equivariant attention mechanism to learn the most representative directional vector for each point and then computes the Cosine distance between each point and its corresponding direction.
To further reduce the information loss caused by rotation-invariant feature extraction, we extend the computation of these two distances to the spatial neighborhood.
This significantly enhances the discriminability of the descriptors while ensuring computational efficiency.

Experimental results show that our method significantly outperforms other rotation-invariant baselines in dealing with rotation issues.
At the same time, our method achieves comparable accuracy to non-rotated point clouds after random rotation in $\mathbb{SO}(3)$, whereas the accuracy of other currently state-of-the-art (SOTA) methods decrease significantly.

Our contributions can be summarized as follows:
\begin{itemize}
    \item We propose a novel place recognition method named VNI-Net, which is a $\mathbb{SO}(3)$ rotation-invariant neural network architecture designed for large-scale LiDAR-based place recognition tasks.
    \item Building upon the rotation-equivariant features extracted by VNN, we design two special rotation-invariant layers based on different distances, and enhance the discriminative power of global descriptors through neighborhood features.
    \item Experimental results demonstrate that our method achieves SOTA performance in addressing rotation issues in place recognition tasks, while maintaining comparable results with current SOTA place recognition methods that do not consider rotation issues.
\end{itemize}

\section{Related Works}

In this section, we first introduce the current state of place recognition.
Then, we discuss methods in computer vision to solve the rotation problem, with a focus on existing 3D point cloud rotation-invariant and rotation-equivariant networks.
Finally, we will summarize the rotation-invariant LiDAR-based place recognition methods.

\subsection{Place Recognition}
The research of place recognition can be divided into vision-based and LiDAR-based methods.
In recent years, vision-based methods have shown good performance  \cite{pengLSDNetLightweightSelfAttentional2022,arandjelovicNetVLADCNNArchitecture,hauslerPatchNetVLADMultiScaleFusion2021,wangTransVPRTransformerbasedPlace2022,geSelfsupervisingFinegrainedRegion2020,benbihiImageBasedPlaceRecognition2020}, but they tend to struggle with challenging illumination variation environments.
On the other hand, LiDAR provides reliable depth information and can maintain excellent performance in cases where visual place recognition fails.

Early LiDAR-based place recognition methods relied on local point features  \cite{sipiranHarris3DRobust2011,zhongIntrinsicShapeSignatures,rohlingFastHistogrambasedSimilarity2015,saltiSHOTUniqueSignatures2014}.
However, local features are sensitive to scene variations.
Moreover, the matching of local features exhibits low computational efficiency, rendering them unsuitable for large-scale location recognition tasks.
Global descriptors can provide a more robust depiction of the scene's overall geometric structure which is unaffected by local variations.
PointNetVLAD  \cite{uyPointNetVLADDeepPoint2018} first uses PointNet \cite{charlesPointNetDeepLearning2017} to extract point features and then aggregates point features using NetVLAD  \cite{arandjelovicNetVLADCNNArchitecture} to generate global descriptors.
PCAN  \cite{zhangPCAN3DAttention} introduces a point-based contextual attention module to generate more distinctive descriptors.
Nevertheless, both PointNetVLAD and PCAN fail to effectively utilize the local neighborhood features of point clouds, leading to a limited capability in representing scenes.
To address this issue, LPD-Net  \cite{liuLPDNet3DPoint2019} uses a neighborhood aggregation module based on graph convolutional networks to obtain local features through multiple hand-crafted features and uses NetVLAD to obtain distinctive global descriptors for localization.
However, LPD-Net is computationally expensive because of the complex hand-crafted input features.
Locus  \cite{vidanapathiranaLocusLiDARbasedPlace2021} proposes using second-order pooling to aggregate segment features and spatiotemporal features into global descriptors.
EPC-Net  \cite{huiEfficient3DPoint2021} proposes a lightweight ProxyConv module based on EdgeConv  \cite{wangDynamicGraphCNN2019} for neighborhood feature aggregation and generates global descriptors using G-VLAD   \cite{huiEfficient3DPoint2021}, which can obtain discriminative scene representations with fewer parameters.
To effectively leverage the sparse characteristics of 3D point clouds, MinkLoc3D  \cite{komorowskiMinkLocLidarMonocular2021} uses 3D sparse convolutions to voxelize sparse point clouds and extract local features using GeM Pooling  \cite{radenovicFineTuningCNNImage2019} instead of NetVLAD.
SVT-Net  \cite{fanSVTNetSuperLightWeight2021} applies self-attention mechanisms in 3D sparse convolutions to learn short-distance local features and long-distance context features using Transformers.

While these methods achieve excellent performance in place recognition, they often struggle when query point clouds undergo rotation, leading to failures in place recognition tasks.

\subsection{Rotation-Invariant convolution}
In 3D point cloud classification tasks, a feature descriptor that possess a strong discriminative ability while exhibiting properties of permutation invariance and rotation invariance is desired.
However, most of the deep learning-based 3D point cloud feature extraction methods focus on the issue of permutation-invariant rather than rotation-invariant.

PointNet  \cite{charlesPointNetDeepLearning2017}, as the pioneering method, directly applies a neural network to point clouds by extracting point-wise features and utilizing pooling to address the issue of point cloud disorderedness.
While this approach mitigates the impact of rotations to some extent by using T-Net to learn a transform, it does not completely solve the rotation problem.
TFN  \cite{thomasTensorFieldNetworks2018} is among the first graph neural networks that possess both translation and rotation equivariant properties.
\cite{estevesLearningEquivariantRepresentations2018} achieves rotation-equivariant by defining convolutional kernel operators based on spherical correlation and transforming spherical convolution into a frequency domain product through a generalized Fourier transform.
Both methods achieve rotation-equivariant by representing convolution kernels as equivariant functions, such as spherical harmonics or circle harmonics.
Unlike the methods that design special convolution kernels,  \cite{finziGeneralizingConvolutionalNeural2020} proposed the Lie group convolution which extracts rotation-invariant features by elevating 3D point clouds to the group space.
This method relaxes the constraint on convolution kernels, but the dimension of the neural network increases due to the lifting operation.
\cite{shen3DRotationEquivariantQuaternionNeural2020}  represents 3D point clouds with quaternions and extracted rotation-invariant features using a specially designed quaternion network.
However, its specially designed quaternion-based convolution methods and the norm-based rotation-invariant feature extraction module reduce the descriptive capability of the features.
VNN \cite{dengVectorNeuronsGeneral2021} converts the scalar features of the network into vector features to achieve rotation-equivariant and develops an effective method for aggregating rotation-equivariant features to obtain rotation-invariant descriptors.
Furthermore, VNN develops comprehensive foundational modules, enabling seamless integration into various 3D point cloud deep learning architectures without altering the input format of the point cloud.

\subsection{Rotation-Invariant LiDAR-based place recognition}
In recent years, many researchers focus on the problem of rotation-induced failure in LiDAR-based place recognition and design rotation-invariant global descriptors in $\mathbb{SO}(2)$ or $\mathbb{SO}(3)$ to address this issue.
To achieve rotation-invariant in $\mathbb{SO}(2)$, OverlapNet  \cite{chenOverlapNetLoopClosing} constructs depth images using multiple information sources in LiDAR scan and estimates the yaw angle using distance images.
RINet  \cite{liRINetEfficient3D2022} designs a rotation-invariant global descriptor that combines semantic and geometric information and achieves rotation-invariant in yaw direction.
For achieving rotation-invariant in $\mathbb{SO}(3)$,  \cite{zhaoSphereVLADAttentionbasedSignalenhanced2022} projects point clouds onto a sphere and use spherical convolution and spherical harmonics to achieve rotation-invariant local descriptors.
Attention mechanisms are employed to enhance the internal connections of local features.
\cite{linSEEquivariantPoint}
employs SE(3)-equivariant group networks to acquire SE(3)-invariant global descriptors, ensuring robustness against rotation and translation variations.
Nonetheless, these methods rely on a discrete representation of rotations on SO(3) using spherical or icosahedral discretization,
which is not fully rotation-invariant.
RPR-Net \cite{fanRPRNetPointCloudbased2022} composes three different levels of rotation-invariant inputs, \ie{relative distances and angles} from the raw point cloud, in the preprocessing stage and employs a dense convolutional network to extract global descriptors.
However, replacing the original point cloud with hand-craft rotation-invariant information can lead to information loss at the early stage, thereby compromising the discriminative nature of global descriptors.

To address these limitations,
we employ VNN to encode the local neighborhood features of the raw point cloud into vectors in a high-dimensional space.
Then, we obtain rotation-invariant features in the high-dimensional space by using different distances on the spatial neighborhood features, resulting in a rotation-invariant global descriptor.
This approach significantly improves the accuracy of the descriptor while ensuring rotation-invariant.

\section{Preliminaries}
\subsection{Rotation-Equivariant and Rotation-Invariant}
For an input point cloud $\mathcal{P} \in \mathbb{R}^{N \times 3}$ and its encoded feature matrix $V \in \mathbb{R}^{N \times C}$, where N is the number of points.
We define an encoding operation as $G : \mathbb{R}^{N \times C_{l}} \rightarrow \mathbb{R}^{N \times C_{l+1}}$, where $C_0 = 3$, indicating the input point cloud.
For any rotation operation $\mathbf{R}  \in$ $\mathbb{SO}(3)$, if the operation G is rotation-equivariant, then it satisfies:
\begin{equation} \label{rotation-equivariant}
    G(\mathbf{R}(V))  = \mathbf{R} (G(V))
\end{equation}
While if the operation G is rotation-invariant, then it satisfies:
\begin{equation} \label{rotation-invariant}
    G(\mathbf{R} (V))  = G(V)
\end{equation}
\subsection{Vector Neurons Network}
VNN  \cite{dengVectorNeuronsGeneral2021} implements a strictly rotation-equivariant network architecture by extending scalar features into vector features and using specially designed activation functions, regularization, and pooling layers.
Specifically, for an input point cloud $\mathcal{P} \in \mathbb{R}^{N \times 3}$, the encoded feature matrix Vector Neurons network is represented as  $V_{vector} \in \mathbb{R}^{N \times C \times 3}$.

The linear layer of VNN uses a bias-free MLP: $f_{lin}(V_l) = WV_l$, where $W \in \mathbb{R}^{C_{l+1} \times C_{l}}$ the weight matrix, $V_l \in \mathbb{R}^{C_{l} \times 3}$ the input, and $V_l \in \mathbb{R}^{C_{l+1} \times 3}$ the output.
Thanks to vector features, this linear mapping is equivalent to any rotation R in $\mathbb{SO}(3)$:
\begin{equation} \label{linear-equivariant}
    f_{lin}(V_l \mathbf{R};W) = WV_l \mathbf{R} = f_{lin}(V_l;W)\mathbf{R} = V_{l+1} \mathbf{R}
\end{equation}
Non-linear layers are essential for improving the representational power of neural networks.
VNN extends the classical ReLU function by dividing vector features into two half-spaces along a certain direction.
Specifically, given an input feature set V, for each vector feature in every dimension of each point $v \in V \in \mathbb{R}^{1 \times 3}$.
VNN learns feature mapping $q \in  \mathbb{R}^{1 \times 3}$ and direction mapping $k \in \mathbb{R}^{1 \times 3}$ through two different weight matrices $W \in \mathbb{R}^{1 \times C}$ and $U \in \mathbb{R}^{1 \times C}$:
\begin{equation}
    q = WV,  k = UV
\end{equation}

The VNN-ReLU is defined as:
\begin{equation}
    v^\prime = \left\{
    \begin{array}{lcl}
        q                                                                                       &  & if\langle q, k \rangle \geqslant  0 \\
        q - \langle q,\frac{k}{\parallel k \parallel } \rangle \frac{k}{\parallel k \parallel } &  & otherwise,
    \end{array}
    \right.
\end{equation}
Based on \Refeq{linear-equivariant}, it can be inferred that $q$ and $k$ are rotation-equivariant with respect to the input feature set $V$.
While the dot product operation can cancel out influence of the rotation matrix:
\begin{equation}
    \langle WV\mathbf{R},UV\mathbf{R} \rangle = \langle q \mathbf{R}, k \mathbf{R} \rangle = \langle q,k \rangle
\end{equation}
Thus, it can be proven that the entire non-linear layer is rotation-equivariant.

\begin{figure*}
    \centering
    \includegraphics[width = \textwidth]{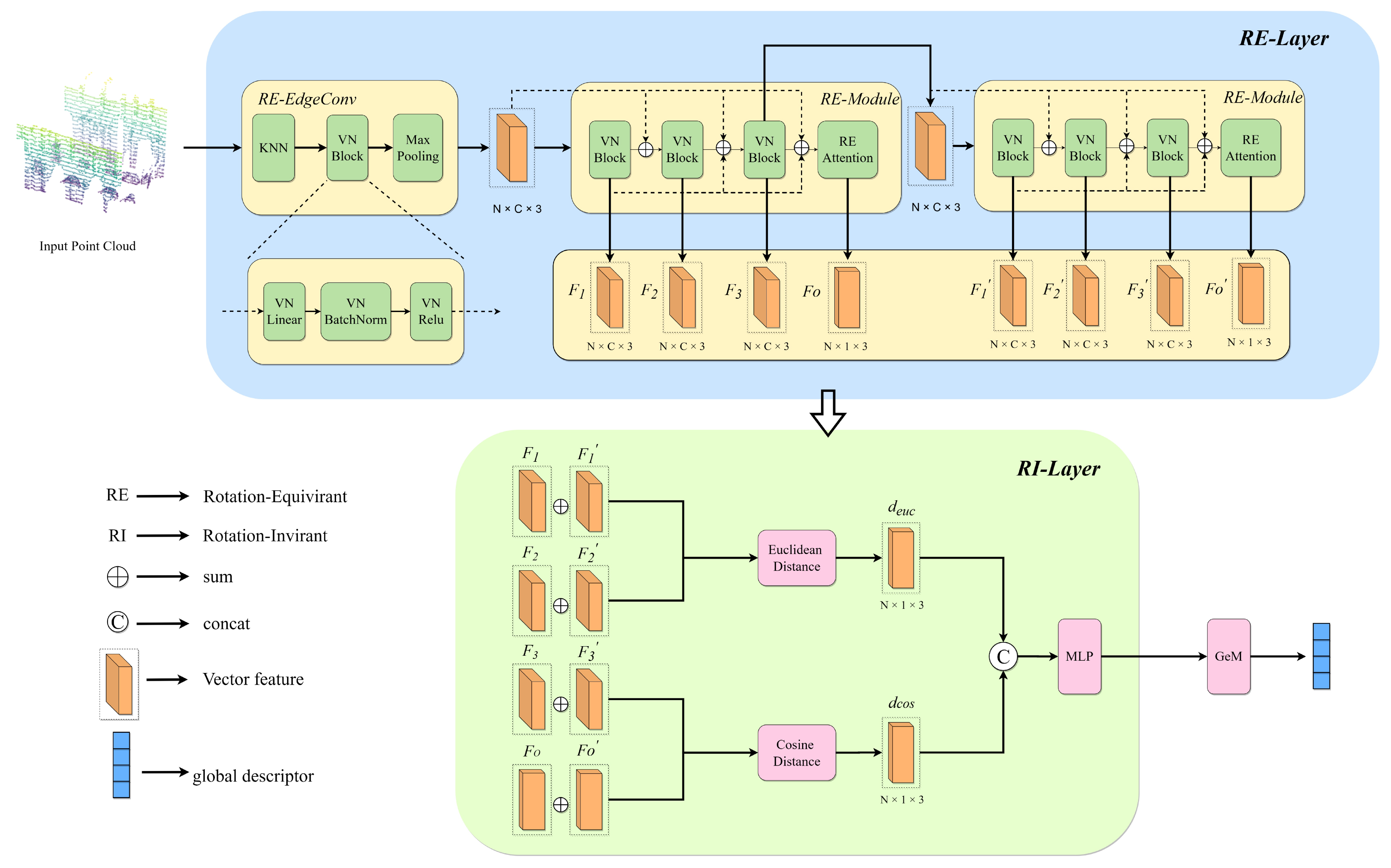}
    \caption{This is the workflow diagram of our proposed method, VNI-Net.
        For the original input point cloud, the process begins with using the RE-EdgeConv to extract rotation-equivariant features within the neighborhood.
        Subsequently, these neighborhood rotation-equivariant features are mapped to different high-dimensional spaces through two RE-Modules, while preserving their rotation-equivariant characteristics.
        Next, a distance-based RI-Layer is employed to compute rotation-invariant features, resulting in a rotation-invariant representation of the point cloud. Finally, the global descriptor is obtained through MLP and GeM layers.
    }
    \label{pipline}
\end{figure*}

\section{Methodology}

In this section, we provide a comprehensive description of VNI-Net, a fully rotation-invariant LiDAR place recognition method based on VNN.
In \Refsec{sec_method_overall}, we begin by introducing the complete network architecture of VNI-Net.
Then detailed descriptions of the key modules are given.
Next, in \Refsec{sec_method_rotation-equivariant},
we describe the dense rotation-equivariant feature extraction layer which obtains multiple distinct rotation-equivariant features.
These features are then used by the distance-based rotation-invariant layers detailed in \Refsec{sec_method_rotation-invariant}, which extracts rotation-invariant feature descriptors.
Finally, in \Refsec{sec_method_loss_function}, we present the loss function along with the training method.

\subsection{The architecture}
\label{sec_method_overall}

The proposed dense and fully rotation-invariant network, namely VNI-Net, is illustrated in \Reffig{pipline}.
We first apply the rotation-equivariant EdgeConv to map the original point cloud to a rotation-equivariant feature space.
Then, the features are fed into two densely connected rotation-equivariant feature extraction modules (RE-Module).
A rotation-invariant layer is used to calculate Euclidean and Cosine distances in different rotation-equivariant feature spaces.
We concatenate these two distances as the rotation-invariant feature output.
Afterwards, we use a standard shared MLP to map each point's features to a higher dimensional feature space.
Finally, GeM \cite{radenovicFineTuningCNNImage2019}, a trainable pooling layer, is adopted to aggregate local features and obtain the rotation-invariant global descriptor.

\subsection{Rotation-Equivariant layer}
\label{sec_method_rotation-equivariant}

\textbf{RE-EdgeConv}
To solve the issue of linear correlation between the rotation-equivariant feature space and the input point cloud coordinates, VNN introduces EdgeConv in the input layer to map $\mathbb{R}^{N \times 1 \times 3}$ to $\mathbb{R}^{N \times C \times 3}(C>1)$.
The EdgeConv constructs local features with $C=3$ in $k$-nearest neighbors as:
\begin{equation}
    f_{local} = (x_i,x_i-x_j, cross(x_i,x_j))
\end{equation}
where $x_j$ is the $j$-th nearest neighbor of $x_i$.
Among these, the coordinates of a point and the differences between the point and its neighbors are rotation-equivariant, denoted as: $x_i\mathbf{R} - x_j\mathbf{R} = (x_i-x_j)\mathbf{R}$.
The cross product is also rotation-equivariant, \ie $cross(x_i\mathbf{R},x_j\mathbf{R}) = cross(x_i,x_j)\mathbf{R}$.
Therefore, by using local features, VNN can obtain rotation-equivariant features with strong descriptive power.

To better describe the local neighborhood information of point clouds in large-scale place recognition tasks, we add local neighborhood centroid to improve the accuracy of local feature description:
\begin{equation}
    f_{local}^\prime = \{x_i,x_i-x_j, cross(x_i,x_j),x_c-x_j,x_i-x_c\}
\end{equation}
where $x_c$ is the centroid of the point cloud formed by the local k-nearest neighbors of point $x_i$.
It can be easily proven that these features are also rotation-equivariant.

Finally, we utilize a vector neural block (VN-Block) that incorporates one-dimensional convolutions, Activation Function, and Batch Normalization to map local features to a high-dimensional space and employ MaxPooling to extract rotation-equivariant features for each point.
The aforementioned features extracted by RE-EdgeConv ensure the rotation-equivariant of the point cloud and effectively represent the local information of the point cloud.

\textbf{RE-Module}
Building upon this, we design a rotation-equivariant convolution module composed of multiple dense connected VN Blocks and RE-Attention.
The first two VN-blocks map the input features to separate rotation-equivariant feature spaces, $F_1$ and $F_2$.
These two features are then added and input into a new VNN block to map to another feature space, $F_3$.

Subsequently, we introduce a Rotation-Equivariant Attention Block to learn the most representative direction vector $F_o$ for each point using \Refeq{ecudistance}.
\begin{equation}
    \label{ecudistance}
    RE_{attention}(Q,K,V) = Softmax(\frac{QK^T}{\sqrt{3C}})V
\end{equation}
where the query map $Q^{N \times C \times 3}$, the key map $K \in \mathbb{R}^{N \times C \times 3}$, and the value map $V \in \mathbb{R}^{N \times C \times 3}$ are obtained by independent VN-linear modules.
Here, $N$ represents the number of points, and $C$ denotes the feature dimension.
Since the input feature $F_3$ exhibits rotation equivariance, after passing through the VN-linear transformation, we have
\begin{equation}
    (QR)(KR)^T = QRR^TK^T = QRR^{-1}K^T = QK^T
\end{equation}
This indicates that our attention module also preserves rotation equivariance.



\subsection{Rotation-Invariant layers}
\label{sec_method_rotation-invariant}

\begin{figure}
    \centering
    \includegraphics[width = \textwidth/2]{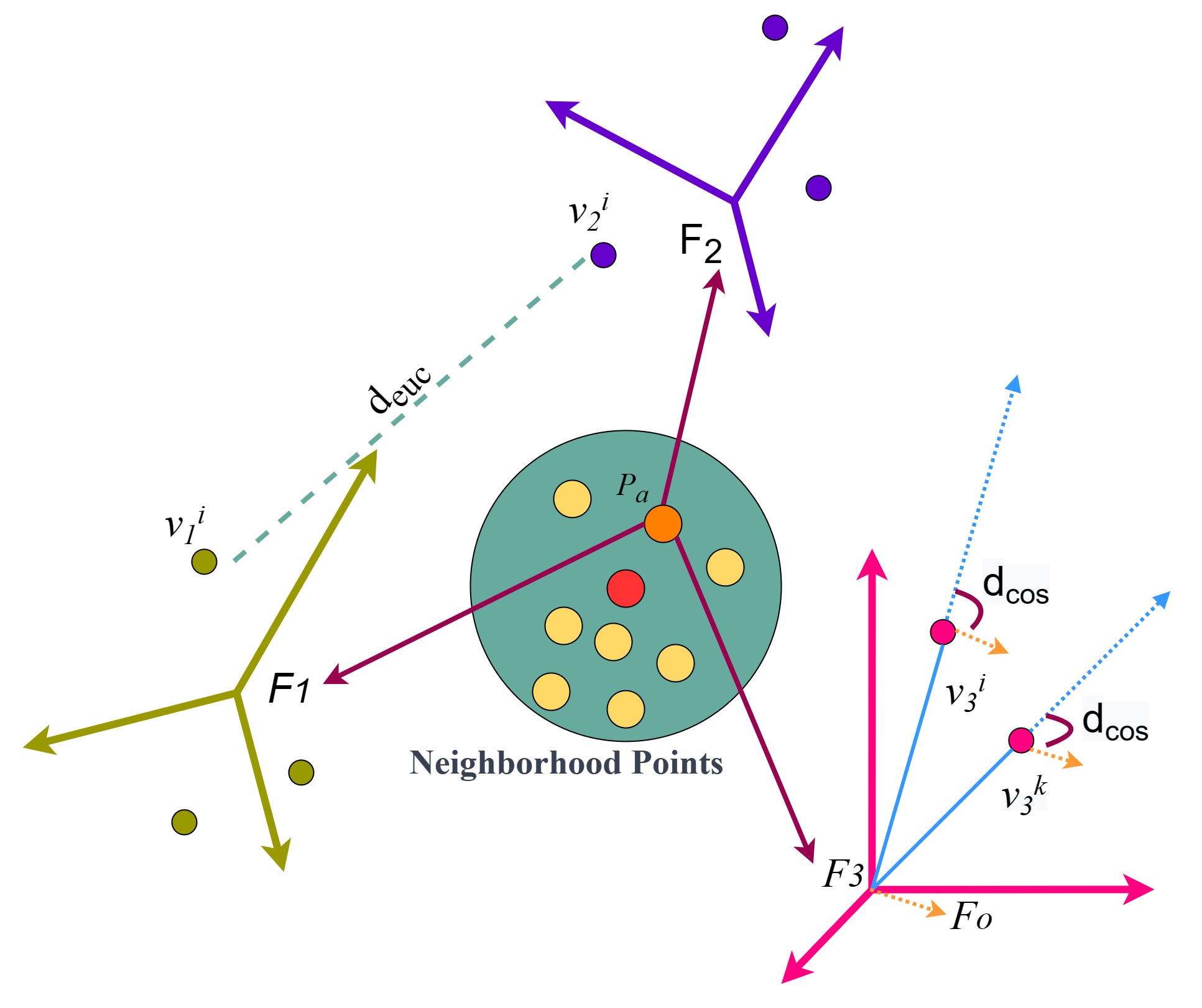}
    \caption{
        The schematic diagram of our proposed specialized distances.
        For each individual point, we begin by mapping the input features to different high-dimensional vector feature spaces, $F_1, F_2$, and $F_3$, using distinct VN-Blocks.
        For the $F_1$ and $F_2$ spaces, we compute the Euclidean distance between the corresponding dimensions of the features.
        For the $F_3$ feature space, we calculate the Cosine distance between each feature vector and the most representative orientation $F_o$ of this point.}
    \label{RIlayer}
\end{figure}

To obtain rotation-invariant features, VNN generates a coordinate system $T \in \mathbb{R}^{3 \times 3}$ from the rotation-equivariant features $V \in \mathbb{R}^{N \times C \times 3}$ using VN-MLP, and obtains rotation-invariant features by reading the rotation-equivariant features $V$ in $T$.
However, this is limited to 3D object classification tasks where the ground truth orientation is assumed to be known.
Inspired by RIConv \cite{zhangRotationInvariantConvolutions2019} in RPR-Net \cite{fanRPRNetPointCloudbased2022}, which constructs rotation-invariant features from relative distances between raw points.
We design rotation-invariant feature extraction layers based on Euclidean distance and Cosine distance in high-dimensional rotation-equivalent feature spaces, as illustrated in \Reffig{RIlayer}.

\textbf{Euclidean distance.}
In the Euclidean space, the distance between two points is not affected by rotation, making it rotation-invariant.
Therefore, we can obtain rotation-invariant features from the Euclidean distance in the rotation-equivariant feature space.
However, to address the issue of decreased discriminability caused by using distance-based descriptors, we calculate the distance in different feature spaces, as shown by $d_{euc}$ in \Reffig{RIlayer}.
\begin{equation}\label{ecudistance}
    d_{euc}^i = \sqrt{(v_1^i - v_2^i)^ 2 }
\end{equation}
where $v_1^i \in F_1 \in \mathbb{R}^{3} $ and $v_2^i \in F_2 \in \mathbb{R}^{3}$ are the $i$-th vector feature in two different feature spaces.
The resulting  $d_{euc} \in \mathbb{R}^{C \times 1}$ is a rotation-invariant feature descriptor that possesses strong representation capabilities.

\textbf{Cosine distance.}
We employ the RE-Attention module to identify the most representative orientation $F_o$ for each point.
Then, we calculate the Cosine distance between each feature vector in the high-dimensional rotation-equivariant feature space and this orientation, as shown by $d_{cos}$ in \Reffig{RIlayer}.
These distances serve as rotation-invariant feature descriptors.
\Refeq{cosdistance} and \Refeq{rotation orthogonality} show that the Cosine distance between vectors in the rotation-equivariant space is also rotation-invariant.

\begin{equation}\label{cosdistance}
    d_{cos} = 1 - \frac{F_3 \cdot F_o}{\parallel F_3 \parallel  \parallel F_o \parallel } = 1 - \frac{F_3 \times (F_o)^T}{\parallel F_3 \parallel  \parallel F_o \parallel }
\end{equation}
where $F_3 \in \mathbb{R}^{C \times 3}$ is the rotation-equivariant feature of a point, while $F_o \in \mathbb{R}^{1 \times 3}$ represents the orientation of the point learned through self-attention.
Since both $F_3$ and $F_o$  are rotation-equivariant, we can infer that:
\begin{equation}
    \label{rotation orthogonality}
    F_3R\times(F_oR)^T=F_3R\times R^TF_o^T=F_3 \times F_o^T
\end{equation}
Finally, we concatenate these two types of rotation-invariant features to obtain $ f_{inv} \in \mathbb{R}^{2C \times 1} $, which serves as the final output of the rotation-invariant layer.
\begin{equation}
    f_{inv} = d_{ecu}\oplus d_{cos}
\end{equation}

\textbf{Neighborhood Aggregation.}
To further improve the discriminability of rotation-invariant descriptors, we calculate the relative distances within each point's neighborhood and use max pooling to obtain the most representative feature description.
Specifically, for each point, we construct a neighborhood point set $\mathcal{H}_i$ in Euclidean space, and then calculate the rotation-invariant feature $f_{inv}$ between point $x_i $ and points $x_j \in \mathcal{H}_i$, and finally take the maximum value of each feature dimension.
\begin{equation}
    \label{neighbor_points}
    f_{inv}^{\prime} = \max\limits_{x_j \epsilon \mathcal{H}_i}(f_{inv}^{ij})=\max\limits_{x_j \epsilon \mathcal{H}_i}(d_{euc}^{ij}\oplus d_{cos}^{ij})
\end{equation}

\subsection{Loss Function and Data Mining}
\label{sec_method_loss_function}

\textbf{Loss Function.}
The network is trained using triplet loss \cite{uyPointNetVLADDeepPoint2018}.
Its objective is to minimize the distance between positive samples,
while maximizing the distance between negative samples.
\begin{equation}
    \mathcal{L} = \sum_i \{d(a_i,p_i) - d(a_i,n_i) + m,0\}
\end{equation}
where $d(x,y) =\| x - y \Vert_2^2 $, and $m$ is a positive constant utilized as a margin to avoid training positive and negative samples to have highly similar features.
To select positive and negative samples, we employed a batch-based positive and negative sample mining strategy in MinkLoc3D \cite{komorowskiMinkLocLidarMonocular2021}, which significantly enhances the training efficiency of the network.
Furthermore, we utilized a hard negative mining strategy to minimize the number of non-zero loss samples during training.

\section{Experiments}

In this section, we first introduce the dataset used in our experiments and provide detailed information on the experimental setup and training strategies.
Subsequently, we compare our experimental results with the SOTA methods and derive conclusion remarks.
We also design specific experiments to evaluate the rotation invariance of our method, demonstrating that our approach is fully rotation-invariant and achieves comparable performance to existing methods that do not account for rotation.
Finally, we conduct ablation experiments to demonstrate the superiority of our proposed method in improving the rotation-invariant global descriptor.

\subsection{Dataset}
The proposed method is evaluated using the benchmark dataset proposed by PointNetVLAD \cite{uyPointNetVLADDeepPoint2018},
which consists of the Oxford RoboCar dataset and three in-house datasets (university sector (U.S.), residential area (R.A.), and business district (B.D.)).
The Oxford RoboCar dataset is constructed by stitching multiple SICK 2D LiDAR scans that cover a range of 20 meters.
While the in-house datasets are collected using the Velodyne HDL 64 3D LiDAR.
Each point cloud frame is associated with a UTM coordinate label.
To facilitate the training,  \cite{uyPointNetVLADDeepPoint2018} preprocesses the dataset by removing ground point clouds, down-sampling the point clouds to a count of 4096, and normalizing the point clouds to a range of [-1,1] to eliminate the impact of translation on accuracy.
The details of the data partition are outlined in \Reftab{dataset_view}.

During training, point cloud frames within a range of 10 meters are defined as positive samples, while those exceeding a range of 50 meters are defined as negative samples.
In the test phase, a correct match is determined when the distance between the retrieved point cloud descriptor and the query falls within a range of 25 meters.
\begin{table}[h]
    \centering
    \small
    \renewcommand{\arraystretch}{1.3}
    \caption{DATASETS SPLITTING}
    \label{dataset_view}
    \begin{tabular}{cccccc}
        \toprule
        \multicolumn{2}{c}{\multirow{2}*{Dataset}} & \multicolumn{2}{c}{Baseline} & \multicolumn{2}{c}{Refine}                   \\
        \multicolumn{2}{c}{}                       & Training                     & Test                       & Training & Test \\
        \hline
        \multicolumn{2}{c}{Oxford}                 & 21.7k                        & 3.0k                       & 21.7k    & 3.0k \\
        \multicolumn{2}{c}{In-house}               & -                            & 4.5k                       & 6.7k     & 1.7k \\
        \bottomrule
    \end{tabular}
    \vspace{-0.3cm}
\end{table}

\begin{figure*}[htbp]
    \centering
    \begin{minipage}[b]{0.5\linewidth}
        \subfigure[Different Network architecture]{
            \label{Network architecture}
            \centering
            \includegraphics[width=0.45\textwidth]{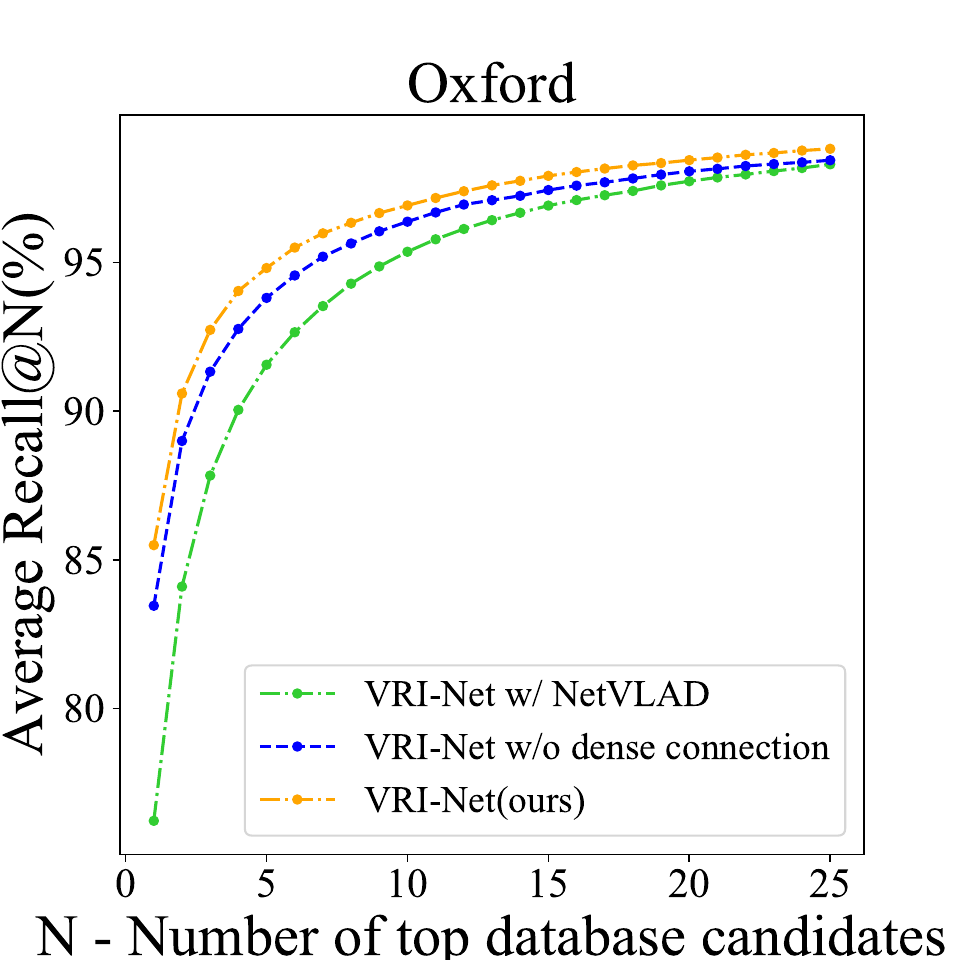}
            \includegraphics[width=0.45\textwidth]{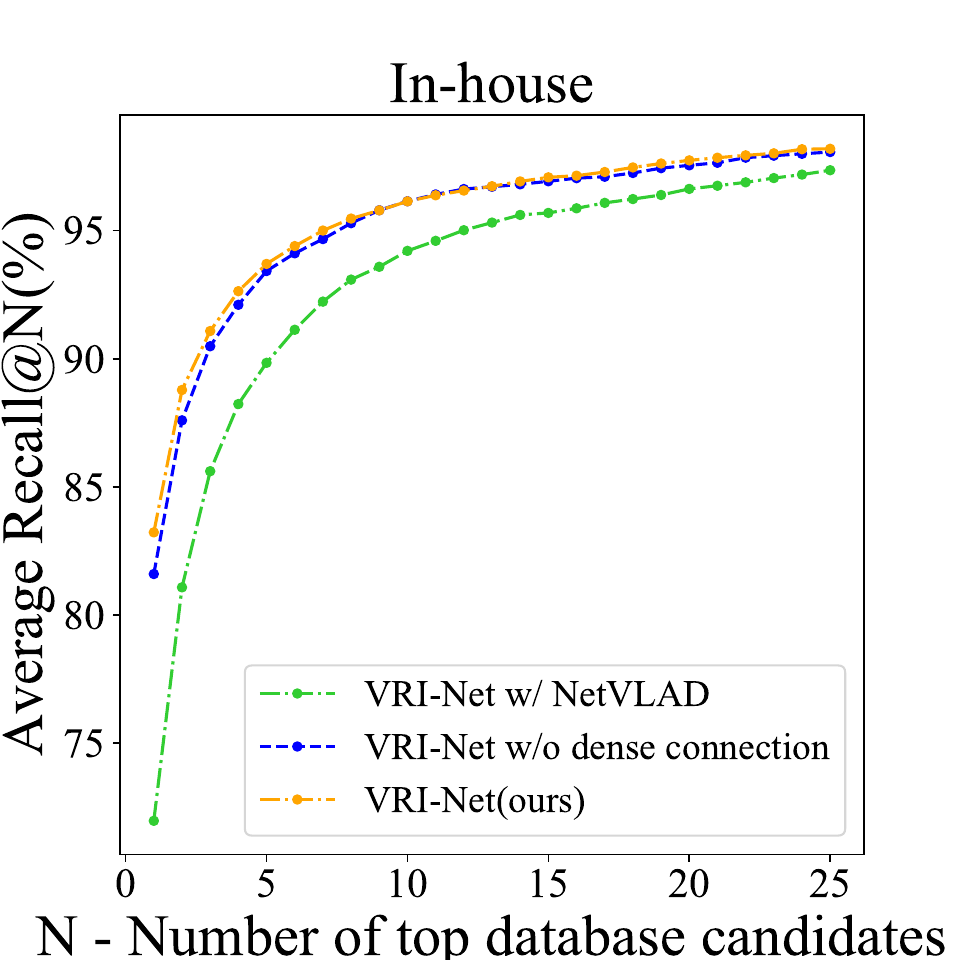}
        }
    \end{minipage}%
    \begin{minipage}[b]{0.5\linewidth}

        \subfigure[Different number of Rotation-Equivariant layers]{
            \label{RELayers}
            \centering
            \includegraphics[width=0.45\textwidth]{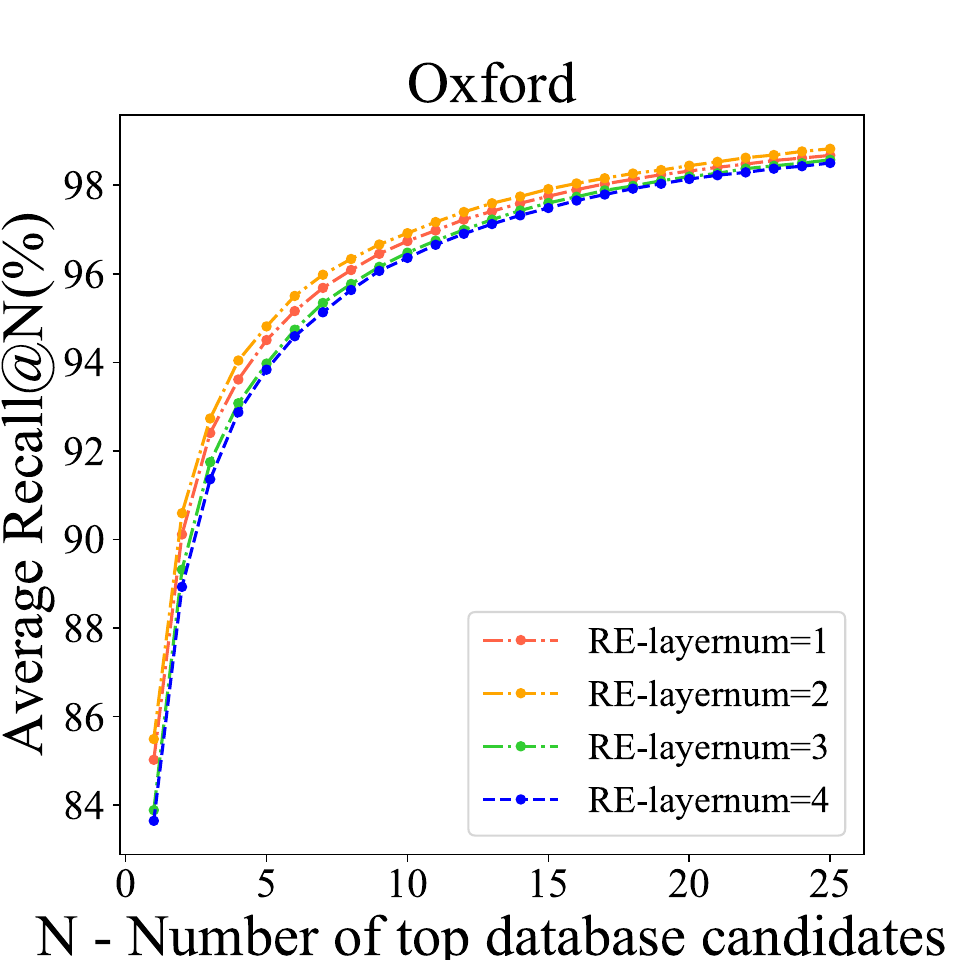}
            \includegraphics[width=0.45\textwidth]{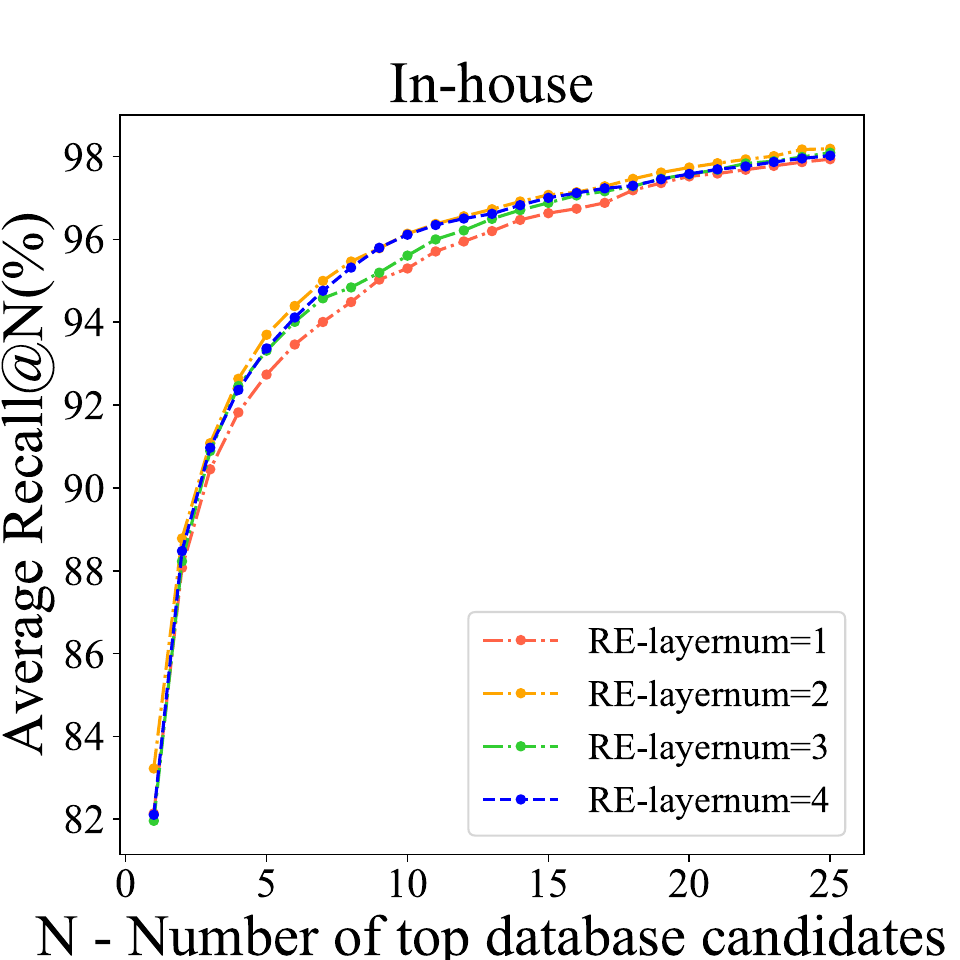}
        }%
    \end{minipage}%
    \caption{
        Ablation study of different Network architectures and different number of Rotation-Equivariant layers trained on the baseline datasets}
\end{figure*}


\subsection{Implement Details and Metrics}
Our experiments are conducted on two NVIDIA 3090 GPUs.
We use the ADAM optimizer to update the network weights with an initial learning rate of 0.01.
We employ a step decay strategy where the learning rate is decreased by a factor of 0.1 at the 20th and 30th epochs.
In order to maintain consistency with the experiments we are comparing against, we adopt two different training settings: baseline and refined.
The baseline setting involves training the network for 40 epochs using only the Oxford RoboCar dataset as the training dataset.
In the refined setting, we add data from the U.S. and B.D. datasets to the training dataset and train the network for 80 epochs.

The dimensionality of our rotation-equivariant features is set to 64, and the rotation-invariant layer concatenates two relative distances from different feature spaces, resulting in a dimensionality of 128.
Our fully connected layers consist of three MLP layers, producing local rotation-equivariant features of dimensionality 1024.
Finally, the GeM Pooling layer outputs a global descriptor of dimensionality 256.
We constructed neighborhood point sets with a neighborhood size of 20.
Furthermore, our triplet loss margin is set to 0.5.

It's worth noting that since our method is fully rotation-equivariant, we do not require any data augmentation techniques during training.
We evaluate our method using the average recall at top 1 (AR@1) and average recall at 1\% (AR@1\%) as performance metrics.

To further validate the robustness of our method against arbitrary rotations, we follow the experimental setup of previous studies and randomly rotated the test point clouds to demonstrate the full rotation invariance capability of our method.

\subsection{Results and discussion}
In this section,
we first quantitatively analyze the performance of global descriptors when point clouds do not undergo any rotation.
Subsequently, we perform random rotations of the point clouds around the z-axis (in $\mathbb{SO}(2)$) and in $\mathbb{SO}(3)$, respectively, to verify the robustness of our method to rotations.
Additionally, we visually inspect the extracted features and demonstrate VNI-Net's rotation invariance.
Finally, we conduct a network complexity analysis.

\begin{table*}[tbp]
    \small
    \renewcommand{\arraystretch}{1.3}
    \begin{center}
        \caption{Quantitative results on Oxford RobotCar Trained on the baseline dataset while evaluated without random rotation}
        \label{tb_worotation}
        \begin{tabular}{lcccccccc}
            \toprule
            \multicolumn{1}{c}{\multirow{2}{*}{Methods}}         & \multicolumn{2}{c}{Oxford} & \multicolumn{2}{c}{U.S.} & \multicolumn{2}{c}{R.A.} & \multicolumn{2}{c}{B.D.}                                                                 \\
                                                                 & AR@1\%                     & AR@1                     & AR@1\%                   & AR@1                     & AR@1\%        & AR@1          & AR@1\%        & AR@1          \\
            \hline
            PointNetVLAD  \cite{uyPointNetVLADDeepPoint2018}     & 80.6                       & 62.8                     & 72.3                     & 63.0                     & 60.4          & 56.7          & 65.7          & 57.4          \\
            PCAN  \cite{zhangPCAN3DAttention}                    & 83.7                       & 69.1                     & 78.9                     & 62.2                     & 71.5          & 56.8          & 66.9          & 58.3          \\
            LPD-Net  \cite{liuLPDNet3DPoint2019}                 & 94.9                       & 86.3                     & 96.2                     & 87.0                     & 90.3          & 82.9          & 89.1          & 82.4          \\
            Minkloc3D \cite{komorowskiMinkLocLidarMonocular2021} & 97.7                       & 93.5                     & 95.0                     & 86.1                     & 91.2          & 80.9          & 88.3          & 82.0          \\
            SVT-Net \cite{fanSVTNetSuperLightWeight2021}         & \textbf{97.8}              & \textbf{93.7}            & 96.6                     & 90.0                     & \textbf{92.7} & 84.3          & \textbf{90.6} & \textbf{85.3} \\
            EPC-Net \cite{huiEfficient3DPoint2021}               & 97.2                       & 92.2                     & \textbf{97.9}            & \textbf{91.4}            & 92.3          & \textbf{85.9} & 88.8          & 84.5          \\
            \hline
            RPR-Net \cite{fanRPRNetPointCloudbased2022}          & 92.2                       & 81.0                     & 94.5                     & 83.2                     & 91.3          & 83.3          & 86.4          & 80.4          \\
            VNI-Net(ours)                                        & \textbf{94.0}              & \textbf{85.5}            & \textbf{95.0}            & \textbf{85.3}            & \textbf{91.5} & \textbf{83.3} & \textbf{86.8} & \textbf{81.4} \\
            \hline
        \end{tabular}
    \end{center}
    \vspace{-0.3cm}
\end{table*}

\textbf{Quantitative results without Rotation.}
We compare our model with several existing methods, including PointNetVLAD \cite{uyPointNetVLADDeepPoint2018}, PCAN \cite{zhangPCAN3DAttention}, LPD-Net \cite{liuLPDNet3DPoint2019}, MinkLoc3D \cite{komorowskiMinkLocLidarMonocular2021}, SVT-Net \cite{fanSVTNetSuperLightWeight2021}, and EPCNet \cite{huiEfficient3DPoint2021}.
For EPCNet, we utilize GeM Pooling \cite{radenovicFineTuningCNNImage2019} for extracting global descriptors, which achieves superior results than those reported in the original paper.
\Reftab{tb_worotation} presents the recall@1 and recall@1\% of all methods on the baseline dataset.
Although there is a slight performance gap between our method and the SOTA methods, it remains competitive.
Furthermore, our method demonstrates strong generalization capabilities on unseen datasets.
This suggests that our designed network architecture, rotation-equivariant feature extraction module, and rotation-invariant module effectively capture both the geometric and semantic features of the original point cloud.
It is important to emphasize that our method does not utilize any data augmentation techniques to enhance performance, and the global descriptors we obtain are fully rotation-invariant.

Compared to the rotation-invariant baseline RPRNet \cite{fanRPRNetPointCloudbased2022}, we observe significant improvements.
This can be attributed to our extraction of rotation-invariant features in the feature space rather than in the original point cloud.

\textbf{Quantitative results with Rotation.}
To validate the rotation invariance of our global descriptors, we first apply random rotations around the z-axis on the original data, simulating the most common rotation challenges in autonomous driving scenarios.
The experimental results on the baseline dataset are shown in \Reftab{tb_rotation}.
Methods such as PointNetVLAD \cite{uyPointNetVLADDeepPoint2018}, PCAN \cite{zhangPCAN3DAttention}, and LPD-Net \cite{liuLPDNet3DPoint2019} attempt to address the rotation problem using the T-Net design in PointNet \cite{charlesPointNetDeepLearning2017}.
Other methods, such as Minkloc3D \cite{komorowskiMinkLocLidarMonocular2021}, SVTNet \cite{fanSVTNetSuperLightWeight2021}, and EPCNet \cite{huiEfficient3DPoint2021} utilize data augmentation to enhance the rotation robustness of global descriptors.
But the experimental results show that these approaches are not effective.

To evaluate our method in more challenging environments, we perform random rotations in $\mathbb{SO}(3)$ on the original point cloud.
The experimental results are shown in \Reftab{tb_rotation}.
The performance of methods that use T-Net or data augmentation to address the rotation problem further declines.
However, our proposed method and the rotation-invariant baseline RPR-Net \cite{fanRPRNetPointCloudbased2022} remain unchanged.
Moreover, our global descriptors exhibit higher scene descriptiveness compared to the baseline.

\begin{table*}[htbp]
    \small
    \renewcommand{\arraystretch}{1.3}
    \begin{center}
        \caption{Quantitative results on Oxford RobotCar Trained on the baseline dataset while evaluated With Random Rotation}
        \label{tb_rotation}
        \begin{tabular}{l|cccc|cccc}
            \toprule
            \multicolumn{1}{c|}{\multirow{2}{*}{Methods}}        & \multicolumn{4}{c|}{w/ z-axis Rotation} & \multicolumn{4}{c}{w/ $\mathbb{SO}(3)$ Rotation}                                                                                                 \\
                                                                 & Oxford                                  & U.S.                                             & R.A.          & B.D.          & Oxford        & U.S.          & R.A.          & B.D.          \\
            \hline
            PointNetVLAD  \cite{uyPointNetVLADDeepPoint2018}     & 22.6                                    & 17.1                                             & 17.9          & 13.5          & 5.3           & 4.9           & 4.6           & 4.0           \\
            PCAN  \cite{zhangPCAN3DAttention}                    & 21.9                                    & 18.2                                             & 21.7          & 16.4          & 5.4           & 6.0           & 4.1           & 4.7           \\
            LPD-Net  \cite{liuLPDNet3DPoint2019}                 & 34.8                                    & 27.6                                             & 27.4          & 19.8          & 8.3           & 6.7           & 7.2           & 7.1           \\
            Minkloc3D \cite{komorowskiMinkLocLidarMonocular2021} & 54.3                                    & 43.2                                             & 44.1          & 50.3          & 12.3          & 14.4          & 10.0          & 13.9          \\
            SVT-Net \cite{fanSVTNetSuperLightWeight2021}         & 55.2                                    & 47.1                                             & 44.5          & 51.2          & 11.9          & 15.3          & 8.8           & 15.4          \\
            EPC-Net \cite{huiEfficient3DPoint2021}               & 58.6                                    & 42.0                                             & 43.5          & 51.3          & 11.7          & 13.4          & 12.0          & 17.9          \\
            \hline
            RPR-Net \cite{fanRPRNetPointCloudbased2022}          & 92.2                                    & 93.8                                             & 91.4          & 86.3          & 92.2          & 93.8          & 91.4          & 86.3          \\
            VNI-Net(ours)                                        & \textbf{94.0}                           & \textbf{95.0}                                    & \textbf{91.5} & \textbf{86.8} & \textbf{94.0} & \textbf{95.0} & \textbf{91.5} & \textbf{86.8} \\
            \hline
        \end{tabular}
    \end{center}
    \vspace{-0.3cm}
\end{table*}

\begin{figure}

    \begin{minipage}[b]{1\linewidth}
        \subfigure[Input Point Cloud]{
            \centering
            \includegraphics[width=0.33\textwidth]{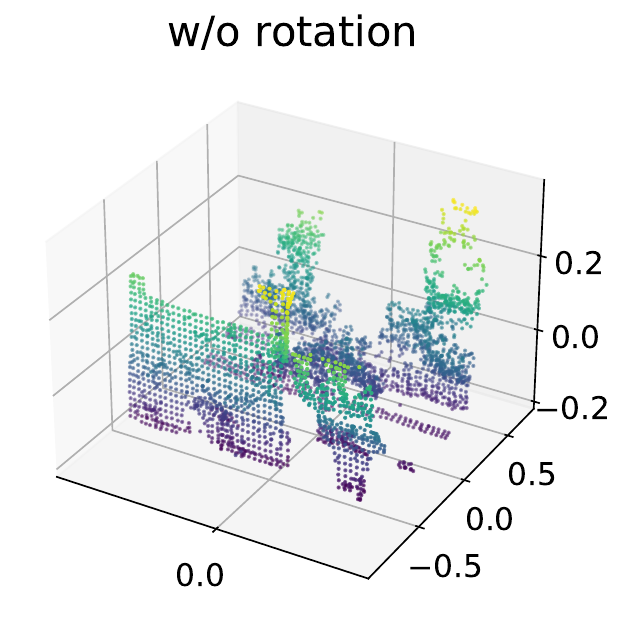}
            \includegraphics[width=0.33\textwidth]{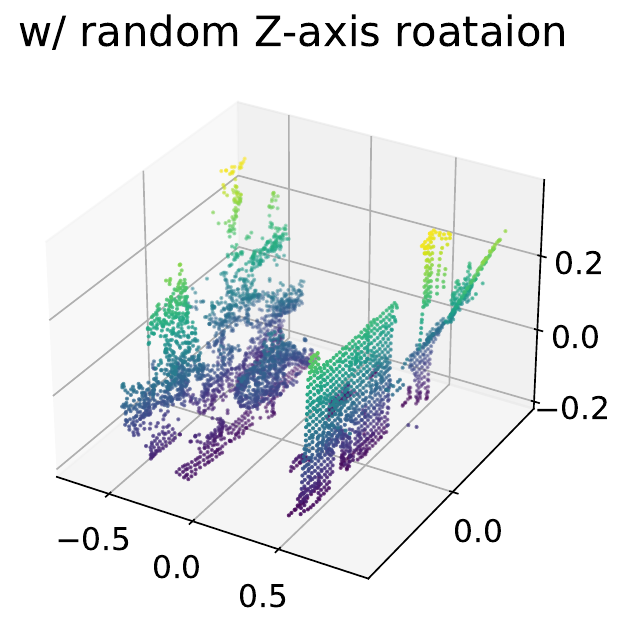}
            \includegraphics[width=0.33\textwidth]{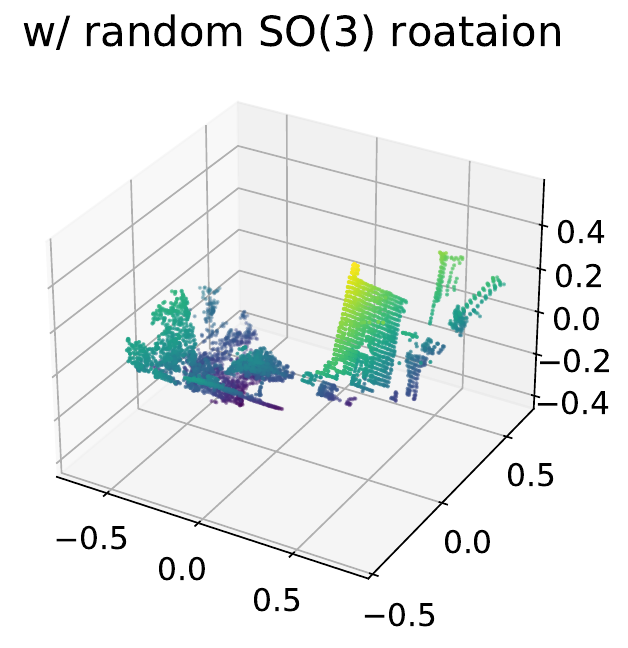}
        }
    \end{minipage}%

    \begin{minipage}[b]{1\linewidth}
        \subfigure[EPCNet local feature]{
            \centering
            \includegraphics[width=0.33\textwidth]{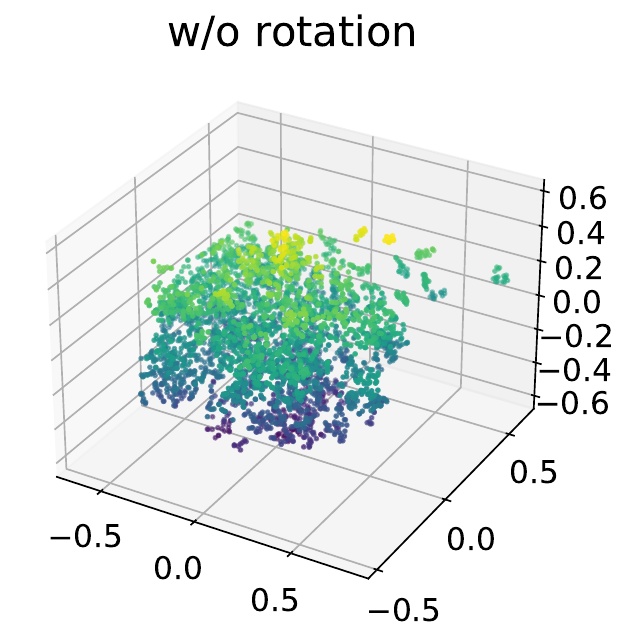}
            \includegraphics[width=0.33\textwidth]{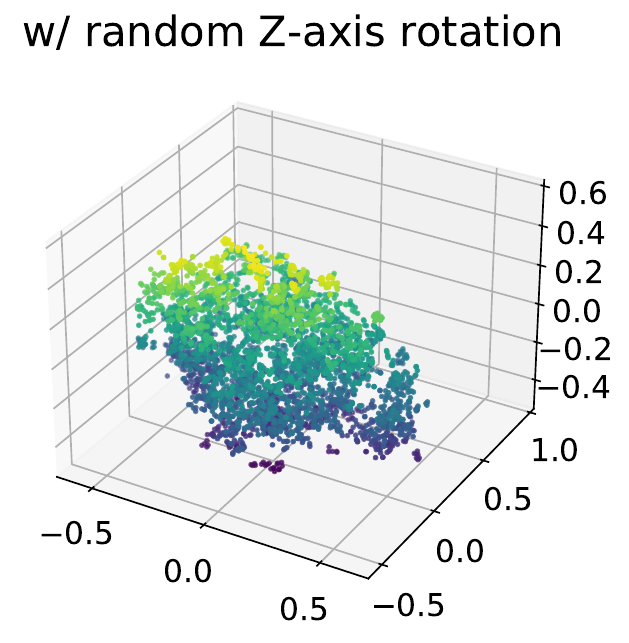}
            \includegraphics[width=0.33\textwidth]{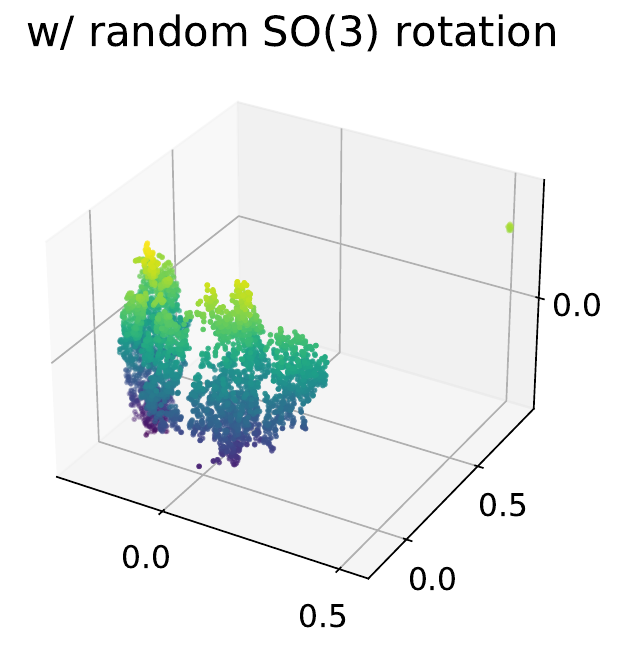}
        }
    \end{minipage}%

    \begin{minipage}[b]{1\linewidth}
        \subfigure[VNI-Net local feature(ours)]{
            \centering
            \includegraphics[width=0.33\textwidth]{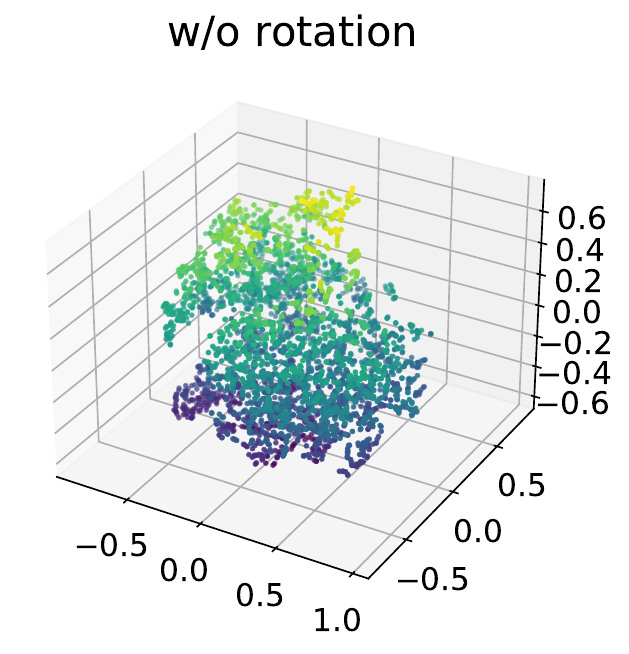}
            \includegraphics[width=0.33\textwidth]{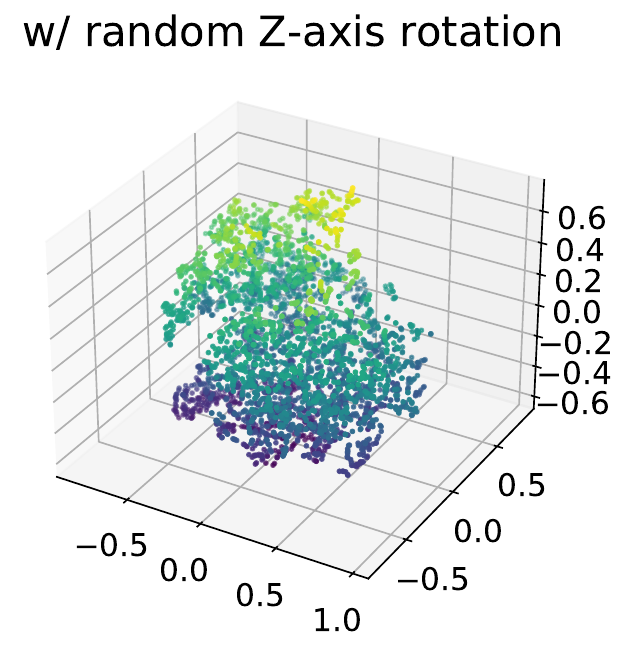}
            \includegraphics[width=0.33\textwidth]{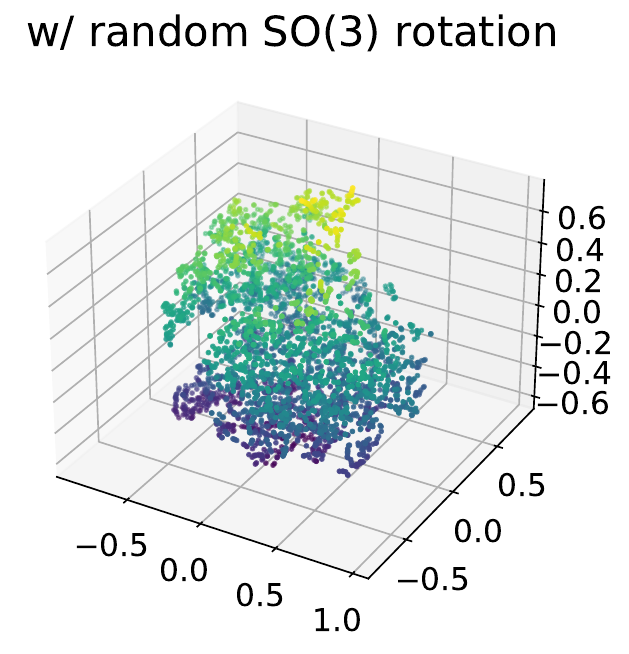}
        }
    \end{minipage}%

    \caption{Visualization of local features Dimensionality reduction by T-SNE. (a) Input PointCloud with different kinds of Rotation. (b) The local features from EPCNet (c) The local features from VNI-Net(ours). It is evident that the local features extracted by EPCNet are affected by rotations. On the other hand, the local features extracted by VNI-Net are entirely rotation-invariant.}
    \label{Qualitative_study_f}
\end{figure}

\textbf{Qualitative analysis.} 
To visually inspect the rotation invariance of our method, we employ T-SNE to visualize the local features extracted from EPCNet \cite{huiEfficient3DPoint2021} and from our method with a randomly rotated input point cloud.
As shown in \Reffig{Qualitative_study_f}, EPCNet, a method that does not consider the rotation invariance of point clouds, is susceptible to the influence of rotations on local features.
In contrast, our model achieves identical feature descriptions across varying rotations, which is attributed to the specially designed rotation-equivariant and rotation-invariant layers.
This visualization demonstrates the effectiveness of our method in capturing and preserving rotation-related information while maintaining rotation invariance in the feature representations.

\textbf{Computational cost.} 
The total number of learnable parameters in our network is 2.2 million, which is twice as many as the current SOTA lightweight network, MinkLoc3D \cite{komorowskiMinkLocLidarMonocular2021} $(1.1M)$ and the rotation-invariant baseline model, RPRNet \cite{fanRPRNetPointCloudbased2022} $(1.1M)$.
But it is still less than other methods such as PointNetVLAD \cite{uyPointNetVLADDeepPoint2018}$(19.8M)$, PCAN \cite{zhangPCAN3DAttention}$(20.4M)$ or PPT-Net \cite{huiPyramidPointCloud}$(13.1M)$.

Our network takes longer to process due to the incorporation of relative distance calculations in the feature space, and the operations to identify the maximum value within the neighborhood.
Specifically, when processing a single frame of point cloud on the 3090 GPU, it takes 0.574 seconds, whereas RPRNet takes 0.238 seconds.

\begin{table}[tbp]
    \small
    \renewcommand{\arraystretch}{1.3}
    \begin{center}
        \caption{Quantitative results on Oxford RobotCar Trained on the refine dataset while evaluated without random rotation}
        \label{refine test}
        \begin{tabular}{lcccc}
            \toprule

            Methods                                              & Oxford        & U.S.          & R.A.          & B.D.          \\
            \hline
            PointNetVLAD  \cite{uyPointNetVLADDeepPoint2018}     & 80.1          & 94.5          & 93.1          & 86.5          \\
            PCAN  \cite{zhangPCAN3DAttention}                    & 86.4          & 94.1          & 92.5          & 87.0          \\
            LPD-Net  \cite{liuLPDNet3DPoint2019}                 & 94.9          & 98.9          & 96.4          & 94.4          \\
            Minkloc3D \cite{komorowskiMinkLocLidarMonocular2021} & \textbf{98.5} & \textbf{99.7} & 99.5          & 95.3          \\
            SVT-Net \cite{fanSVTNetSuperLightWeight2021}         & 98.4          & 99.9          & 99.5          & \textbf{97.2} \\
            EPC-Net \cite{huiEfficient3DPoint2021}               & 97.9          & 99.6          & \textbf{99.7} & 96.3          \\
            \hline
            RPR-Net \cite{fanRPRNetPointCloudbased2022}          & 93.6          & 97.9          & \textbf{97.4} & 92.4          \\
            VNI-Net(ours)                                        & \textbf{94.2} & \textbf{98.3} & 96.8          & \textbf{93.0} \\
            \hline
        \end{tabular}
    \end{center}
    \vspace{-0.3cm}
\end{table}

\subsection{Ablation Study}
In this section, we design ablation experiments to validate the effectiveness of the proposed modules.
All experiments are conducted on the baseline dataset and evaluated on the Oxford and In-house datasets (U.S., R.A., and B.D.).
To simplify the evaluation, for the In-house dataset, we present the final results using the mean values.

\textbf{Different methods for local neighborhood feature extraction.}
\Reftab{ablation 1} presents the recall rates using different EdgeConv configurations.
It can be observed that using the proposed five-dimensional features with EdgeConv yields the best performance.
This is because real-world point clouds often exhibit irregular distributions.
When there is a highly uneven distribution of points within the k-nearest neighbors, introducing neighborhood centroid can provide a more accurate depiction of the point cloud distribution in that neighborhood.

\begin{table}[tbp]
    \small
    \renewcommand{\arraystretch}{1.3}
    \caption{Ablation studies of Different features in EdgeConv}
    \label{ablation 1}
    \begin{tabular}{lcc}
        \toprule
        \multicolumn{1}{c}{\multirow{2}*{Models}}                                             & \multicolumn{1}{c}{Oxford} & \multicolumn{1}{c}{In-house} \\
        \multicolumn{1}{l}{}                                                                  & \multicolumn{1}{c}{AR@1\%} & \multicolumn{1}{c}{AR@1\%}   \\
        \hline
        \multicolumn{1}{l}{RPR-Net \cite{fanRPRNetPointCloudbased2022} }                      & 92.2                       & 90.7                         \\
        \multicolumn{1}{l}{VNI-Net w/o EdgeConv}                                              & -                          & -                            \\
        \multicolumn{1}{l}{VNI-Net w/ \, EdgeConv(dim=3) \cite{dengVectorNeuronsGeneral2021}} & 92.8                       & 89.1                         \\
        \multicolumn{1}{l}{VNI-Net w/ \, EdgeConv(dim=5)}                                     & \textbf{94.0}              & \textbf{91.0}                \\
        \bottomrule
    \end{tabular}
    \vspace{-0.3cm}
\end{table}

\textbf{The design of rotation-invariant layers.}
In this section, we analyze the performance of different components in the rotation-invariant module, the results are shown in
\Reftab{ablation 2}.
It can be seen that the rotation-invariant structure proposed in VNN  \cite{dengVectorNeuronsGeneral2021} performs poorly in large-scale place recognition tasks.
This confirms our statement that finding an accurate absolute coordinate system for real-world 3D point cloud frame is difficult.
We further find that the Euclidean distance extracted in the feature space $d_{euc}$ is identified as the most representative component of the global descriptor.
Removing $d_{euc}$ results in a noticeable decline in both performance and generalization of the network.
Moreover, experiments on the In-house dataset demonstrate that the Cosine distance $d_{cos}$ and the self-attention module used to extract the orientation of each point can significantly enhance the generalization of our method on unseen datasets.
Finally, it is evident that our design of computing distances between different points using K-nearest neighbors and utilizing to identify the most representative rotation-invariant features (Neighborhood in \Reftab{ablation 2}) significantly improves the performance of the global descriptors.
\begin{table}[tbp]
    \small
    \renewcommand{\arraystretch}{1.3}
    \caption{Ablation studies of different Rotation-Invariant layer}
    \label{ablation 2}
    \begin{tabular}{lcc}
        \toprule
        \multicolumn{1}{c}{\multirow{2}*{Models}}                                     & \multicolumn{1}{c}{Oxford} & \multicolumn{1}{c}{In-house} \\
        \multicolumn{1}{l}{}                                                          & \multicolumn{1}{c}{AR@1\%} & \multicolumn{1}{c}{AR@1\%}   \\
        \hline
        \multicolumn{1}{l}{VNI-Net+VNN-INVLayer \cite{dengVectorNeuronsGeneral2021} } & 88.5                       & 86                           \\
        \multicolumn{1}{l}{VNI-Net w/o $d_{euc}$}                                     & 90.6                       & 89.1                         \\
        \multicolumn{1}{l}{VNI-Net w/o $d_{cos}$}                                     & 93.3                       & 89.0                         \\
        \multicolumn{1}{l}{VNI-Net w/o Self-Attention}                                & 93.3                       & 90.7                         \\
        \multicolumn{1}{l}{VNI-Net w/o Neighborhood}                                  & 91.7                       & 88.9                         \\
        \multicolumn{1}{l}{VNI-Net}                                                   & \textbf{94.0}              & \textbf{91.0}                \\
        \bottomrule
    \end{tabular}
    \vspace{-0.3cm}
\end{table}

\textbf{The impact of network architecture.}
From \Reffig{Network architecture}, it can be observed that the densely connected network architecture is highly effective in improving the performance of global descriptors.
By adding the shallow features to the deep features, the network focuses on the original point cloud's geometric structure and semantic information, enhancing the representational power of the descriptors for the scene.
Additionally, we tested the influence of different feature aggregation methods, \ie{NetVLAD, GeM}, on the performance of global descriptors.
The experiments show that GeM Pooling, with fewer parameters, achieves better results.
This can be attributed to the high quality of local features extracted by the neighborhood mechanism and dense network mechanism, which effectively capture the information of the point cloud.
Therefore, using a GeM layer with minimal trainable parameters can aggregate a discriminative global feature.

\textbf{The number of rotation-equivariant layers.}
\Reffig{RELayers} presents the recall curves on the baseline dataset for different numbers of rotation-equivariant layers.
It can be observed that our model achieves the best performance when two rotation-equivariant layers are stacked.
As the number of layers increases, the performance on the in-house dataset noticeably declines, indicating overfitting in the network.

\section{Conclusion}


In this paper, we present VNI-Net, a novel fully rotation-invariant network designed for large-scale LiDAR-based place recognition. VNI-Net is built upon VNN and incorporates rotation-equivariant blocks to effectively map features into a high-dimensional rotation-equivariant space. To overcome the challenge of preserving local spatial information in point clouds while utilizing rotation-invariant descriptors, we propose the rotation-invariant feature extraction module based on two distinct distance metrics. These modules not only ensure the discriminative nature of the descriptors but also enhance the point cloud's resilience to arbitrary rotations under the $\mathbb{SO}(3)$. The effectiveness and rotational robustness of our proposed approach are verified through evaluations conducted on the Oxford dataset and an In-house dataset. In the future, we aim to streamline the distance computation while maintaining rotation invariance in order to further enhance the extraction speed of global descriptors.


\bibliography{reference}
\bibliographystyle{IEEEtran}

\end{document}